\documentclass[conference]{IEEEtran}
\IEEEoverridecommandlockouts
\usepackage{cite}
\usepackage{amsmath,amssymb,amsfonts}
\usepackage{algorithmic}
\usepackage{graphicx}
\usepackage{textcomp}
\usepackage{tabularx}
\usepackage{siunitx}
\usepackage{xcolor}
\usepackage{balance}
\usepackage{subcaption} 

\newtheorem{mydef}{Definition}
\newtheorem{theorem}{Theorem}
\newtheorem{remark}{Remark}

\def\BibTeX{{\rm B\kern-.05em{\sc i\kern-.025em b}\kern-.08em
    T\kern-.1667em\lower.7ex\hbox{E}\kern-.125emX}}

\begin{document}
\title{Obstacle Avoidance for Unicycle-Modelled Mobile Robots with Time-varying Control Barrier Functions}
\author{Jihao Huang$^{1}$, Zhitao Liu$^{1\dagger}$, Jun Zeng$^{2}$, Xuemin Chi$^{1}$, Hongye Su$^{1}$%
\thanks{This work was supported in part by National Key R\&D Program of China (Grant NO. 2021YFB3301000); National Natural Science Foundation of China (NSFC:62173297), Zhejiang Key R\&D Program (Grant NO. 2022C01035), Fundamental Research Funds for the Central Universities (NO.226-2022-00086).}
\thanks{
$^{1}$Jihao Huang, Xuemin Chi, Zhitao Liu and Hongye Su are with the State Key Laboratory of Industrial Control Technology, Institute of Cyber-Systems and Control, Zhejiang University, Hangzhou, China {\tt\footnotesize \{jihaoh, chixuemin, ztliu, hysu\}@zju.edu.cn}.

$^{2} $Jun Zeng is with Cruise LLC, USA {\tt\footnotesize jun.zeng@getcruise.com}.}%
\thanks{$^\dagger$ Corresponding author.}
}

\maketitle
\begin{abstract}
In this paper, we propose a safety-critical controller based on time-varying control barrier functions (CBFs) for a robot with an unicycle model in the continuous-time domain to achieve navigation and dynamic collision avoidance.
Unlike previous works, our proposed approach can control both linear and angular velocity to avoid collision with obstacles, overcoming the limitation of confined control performance due to the lack of control variable. 
To ensure that the robot reaches its destination, we also design a control Lyapunov function (CLF).
Our safety-critical controller is formulated as a quadratic program (QP) optimization problem that incorporates CLF and CBFs as constraints, enabling real-time application for navigation and dynamic collision avoidance. 
Numerical simulations are conducted to verify the effectiveness of our proposed approach.
\end{abstract}


\section{Introduction}
The field of robotics has experienced rapid development, allowing for their application in various areas such as delivery, autonomous driving, and rescue operations~\cite{lee2023survey}.
Therefore, designing a safety-critical controller that can be used for all applications is of great significance.
Safety-critical control, which means the controller prioritizes safety over other things like tracking or planning~\cite{thontepu2022control}.
Ensuring the safety of robot requires the robot to avoid collisions with obstacles.
Researchers have developed many methods to address this issue, including model predictive control (MPC)~\cite{zeng2021safety}, artificial potential fields~\cite{singletary2021comparative}, reachability analysis\cite{li2021comparison} and neural networks~\cite{xiao2023barriernet}.
Recently, control barrier function (CBF) based approaches are widely used in safety-critical control systems~\cite{ames2019control}. 

A safety-critical controller which unifies control Lyapunov function (CLF) for stability and CBF for safety through a quadratic program (CLF-CBF-QP)~\cite{ames2014control, ames2016control} are proposed to achieve adaptive cruise control (ACC) in the continuous time domain.
This approach efficiently addresses real-time practical applications as the safety and stability constraints are affine in the control variable.
Moreover, this paradigm also has been used successfully in robotics for navigation and collision avoidance~\cite{wu2016safety, thontepu2022control, he2021rule} and has been extended into various variant approaches~\cite{zeng2021feasibility, garg2021robust, nguyen2022adaptive}. 

However, implementing a continuous-time CBF based approach for collision avoidance with obstacles for the robot with an unicycle model is challenging, since the control system is non-affine with linear and angular velocities as system inputs.
Moreover, a nominal CBF won't be able to ensure that the system is dependent on all control inputs, i.e., the robot is unable to avoid obstacles by controlling steering, it loses control over the steering.
Some work utilizes High-Order CBF (HOCBF)~\cite{xiao2019control} to address this issue, however, the coupling between linear velocity and angular velocity in HOCBF makes it hard to design an affine constraint on control variables.
To maintain an affine constraint, researchers opt to assume the linear velocity is fixed and only adjust the angular velocity to avoid collisions with obstacles~\cite{xiao2021high}. 
Nevertheless, this approach approximates the system dynamics and loses the fidelity with respect to the original model.
In summary, existing work for collision avoidance based on CBF for the unicycle model in continuous time domain only has limited control performance due to either some assumptions or approximations, which can result in deadlock or even unsafe behaviors due to simulation fidelity gap.

Some studies~\cite{minh2022safety,jian2023dynamic} choose to address the aforementioned issues by utilizing discrete CBF (DCBF) in a discrete-time domain through MPC~\cite{enhancing2022zeng}.
Since the dynamics of the unicycle model is discrete in the discrete time domain, collision avoidance with obstacles can be greatly affected by both linear and angular velocities.
This ultimately leads to optimal control performance.
However, when there are many obstacles in the environment, satisfying real-time requirements may not be feasible in this manner, since MPC problem needs to be linearized locally with respect to the system dynamics~\cite{liu2023iterative}.
Nevertheless, the complexity still increases dramatically with a large number of obstacles.

To solve the problems of limited control performance in the continuous time domain and real-time performance in the discrete time domain, we introduce a novel control barrier function that utilizes the coordinates of the rear axle axis to enable the robot with an unicycle model to achieve collision avoidance in the continuous time domain.
Our formulation handles well both time-varying~\cite{nguyen2016optimal, wang2022high, igarashi2019time} and time-invariant continuous-time CBFs, making it suitable for avoiding collisions with both static and dynamic obstacles, without dramatic complexity even with a large number of obstacles.
With this formulation, the controller is allowed for control over both linear velocity and angular velocity to avoid collision with obstacles.
Additionally, we also design a CLF to ensure the robot converge to its destination regardless of start and goal positions.
We demonstrate the effectiveness of our proposed approach through numerical simulations involving multiple static and dynamic obstacles.

The paper is organized as follows:
We formulate the problem of this work and present the background information about CLF and CBF in Sec.~\ref{sec:back}.
In Sec.~\ref{sec:control_design}, we design the CLF and CBF according the requirement of the control objective and formulate an optimal control problem which incorporates CLF and CBFs as constraints.
To verify the effectiveness of our approach for navigating with obstacles, numerical validations are presented in Sec.~\ref{sec:experiment}. 
Sec.~\ref{sec:conclusion} provides concluding remarks.

\section{Background}
\label{sec:back}
In this section, we provide some relevant background information to formulate our problem.
Firstly, we define the problem and describe the unicycle model of the robot that is considered in this work.
Following this, we introduce two important concepts: control Lyapunov function (CLF) and control barrier function (CBF).

\subsection{Problem Formulation}
In our work, we aim to guide a robot to its destination while ensuring it avoids all static and dynamic obstacles.
Assume the robot is in a circular shape, and the dynamics of the unicycle model for the robot are presented below:
\begin{equation} 
    \left[\begin{array}{c}
    \dot{x}_\text{p} \quad
    \dot{y}_\text{p} \quad
    \dot{\theta}
    \end{array}\right]^T = 
    \left[\begin{array}{c}
    v \cos \theta \quad
    v \sin \theta \quad
    \omega
    \end{array}\right]^T,
    \label{eq:dubins}
\end{equation}
where $x_\text{p}, y_\text{p}, \theta$ denote the current coordinates of the rear axle axis and the orientation of the robot with respect to the $x$ axis, $v, \omega$ denote the linear and the angular velocity.
Moreover, the center of the robot could be represented as follows:
\begin{equation} 
    \left[\begin{array}{c}
    x_\text{c} \quad
    y_\text{c} 
    \end{array}\right]^T = 
    \left[\begin{array}{c}
    x_\text{p} + l \cos \theta \quad
    y_\text{p} + l \sin \theta 
    \end{array}\right]^T
    \label{eq:position}
\end{equation}
where $l$ is the distance between the rear axle axis and the center of the robot.

Many works\cite{wu2016safety, thontepu2022control} based on the unicycle model only consider the center position of the robot, without considering using the rear axle axis as a representation of the center position coordinates, thus leads to the imperfect control performance, more details can refer to Rem.~\ref{remark2}.

\subsection{Control Lyapunov Function (CLF)}
In this section, we will define the control Lyapunov function (CLF), which is commonly utilized to achieve the control objective of stabilizing a system to an equilibrium state.
This concept also naturally leads to the "dual" for safety: control barrier function (CBF).

Suppose that we have a nonlinear affine control system defined as follows:
\begin{equation}
    \dot{\mathbf{x}} = f(\mathbf{x}) + g(\mathbf{x})\mathbf{u} ,
    \label{eq:system}
\end{equation}
where $\mathbf{x} \in \mathbb{R}^n$ and $\mathbf{u} \in \mathbb{R}^m$ denote the state and control of the system, with $f$ and $g$ being locally Lipschitz.
The state of the system satisfies $\mathbf{x} \in \mathcal{D} \subset \mathbb{R}^n$.
The system is also subject to the input constraints
\begin{equation}
    \mathbf{u} \in \mathcal{U} := \{\mathbf{u} \in \mathbb{R}^m, \mathbf{u}_\text{min} \leq \mathbf{u} \leq \mathbf{u}_\text{max} \},
\end{equation}
where $\mathcal{U}$ denotes the set of admissible inputs, $\mathbf{u}_\text{min}$ and $\mathbf{u}_\text{max}$ represent the lower and upper bounds of $\mathbf{u}$.
Before defining the CLF, we first need to introduce the concept of class $\mathcal{K}$ function $\alpha$: 
$[0, a) \to [0, \infty)$ is said to belong to class $\mathcal{K}$ if it is strictly increasing and has $\alpha(0) = 0$.
It is said to belong to class $\mathcal{K}_{\infty}$ if it belongs to class $\mathcal{K}$ function and satisfies $a = \infty$ and $\alpha(b) \to \infty$ as $b \to \infty$.
And an extended class $\mathcal{K}_{\infty}$ function is a function $\alpha$: $\mathbb{R} \to \mathbb{R}$ that is strictly increasing with $\alpha(0) = 0$ and $\alpha(b) \to \infty$ as $b \to \infty$.
\begin{mydef}
    A continuously differential function $V: \mathbb{R}^n \to \mathbb{R}$ is a control Lyapunov function if it is positive definite and satisfies~\cite{ames2019control}:
    \begin{equation}
        \inf_{\mathbf{u} \in \mathcal{U}} [L_f V(\mathbf{x}) + L_g V(\mathbf{x}) \mathbf{u}] \leq -\gamma(V(\mathbf{x})),
        \label{eq:define_clf}
    \end{equation}
    where $L_f V(\mathbf{x})$ and $L_g V(\mathbf{x})$ are Lie-derivatives of $V(\mathbf{x})$, $\gamma$ is a class $\mathcal{K}$ function.
\end{mydef}

We can obtain the set of controls that enable the system to be stable for every $\mathbf{x} \in \mathcal{D}$:
\begin{equation}
    K_{\text{clf}} (\mathbf{x}):=\{\mathbf{u} \in \mathcal{U}, L_f V(\mathbf{x}) + L_g V(\mathbf{x}) \mathbf{u} \leq -\gamma(V(\mathbf{x})) \}.
    \label{eq:kclf}
\end{equation}

\begin{equation}
    L_g V(x) = \frac{\partial V(x)}{\partial x} g
\end{equation}
So we can utilize this affine constraint in $\mathbf{u}$ to formulate an optimization based controller.

\subsection{Control Barrier Function (CBF)}
Unlike CLF, which leads the system to an equilibrium state, CBF is proposed in the context of safety.
We define a set $\mathcal{C}$ as a superlevel set of a continuously differentiable function $h$: $\mathcal{D} \subset \mathbb{R}^n \to \mathbb{R}$, yielding:
\begin{equation}
\begin{aligned}
    \mathcal{C} & = \{ \mathbf{x} \in \mathcal{D} \subset \mathbb{R}^n : h(\mathbf{x}) \geq 0 \}, \\
    \partial \mathcal{C} & = \{ \mathbf{x} \in \mathcal{D} \subset \mathbb{R}^n : h(\mathbf{x}) = 0 \}, \\
    \rm Int(\mathcal{C}) & = \{ \mathbf{x} \in \mathcal{D} \subset \mathbb{R}^n : h(\mathbf{x}) > 0 \}.
    \label{eq:safe_set}
\end{aligned} 
\end{equation}
We refer to $\mathcal{C}$ as the safe set.

\begin{mydef}
    The set $\mathcal{C}$ is forward invariant if for every $\mathbf{x}_0 \in \mathcal{C}, \mathbf{x}(t) \in \mathcal{C}$ for $\mathbf{x}(0) = \mathbf{x}_0, \forall t \geq 0$. 
    The system \eqref{eq:system} is safe with respect to the safe set $\mathcal{C}$ if $\mathcal{C}$ is forward invariant.
\end{mydef}

\begin{mydef}
    Given the safe set $\mathcal{C}$ defined by \eqref{eq:safe_set}, with $\frac{\partial h(\mathbf{x})}{\partial \mathbf{x}} \not=0, \forall \mathbf{x} \in \partial \mathcal{C}$, the function $h$ is called the control barrier function (CBF) defined on the set $\mathcal{D}$, if there exists an extended class $\mathcal{K}_{\infty}$ function $\alpha$ such that the system \eqref{eq:system} satisfies~\cite{ames2019control}:
    \begin{equation}
        \sup_{\mathbf{u} \in \mathcal{U}} [L_f h(\mathbf{x}) + L_g h(\mathbf{x}) \mathbf{u}] \geq -\alpha(h(\mathbf{x})),
        \label{eq:define_cbf}
    \end{equation}
    where $L_f h(\mathbf{x})$ and $L_g h(\mathbf{x})$ are Lie-derivatives of $h(\mathbf{x})$.
\end{mydef}

We can also obtain the set of controls which render $\mathcal{C}$ safe for all $\mathbf{x} \in \mathcal{D}$:
\begin{equation}
    K_{\text{cbf}} (\mathbf{x}):=\{ \mathbf{u} \in \mathcal{U}, L_f h(\mathbf{x}) + L_g h(\mathbf{x}) \mathbf{u} \geq -\alpha(h(\mathbf{x})) \}.
    \label{eq:kcbf}
\end{equation}

\begin{theorem}
    If $h$ is a CBF on $\mathcal{D}$ and $\frac{\partial h(\mathbf{x})}{\partial \mathbf{x}} \not=0, \forall \mathbf{x} \in \partial \mathcal{C}$, then any Lipschitz continuous controller $\mathbf{u}(\mathbf{x}) \in K_{\text{cbf}} (\mathbf{x})$ for the system \eqref{eq:system} can guarantee the forward invariance of the set $\mathcal{C}$ and thus safety.
\end{theorem}

The constraint \eqref{eq:kcbf} in $\mathbf{u}$ is also an affine constraint, so we can combine it with \eqref{eq:kclf} to formulate the optimization based controller.
We also need to pay attention that when the system needs to meet different safety constraints, the safety set of the system is the intersections of the safety sets corresponding to different constraints.
To simplify the problem, we choose to use a constant scalar as the class $\mathcal{K}$ function for both CLF and CBF inequalities, i.e., $\gamma(V(\mathbf{x})) = \gamma V(\mathbf{x})$, $\alpha(h(\mathbf{x})) = \alpha h(\mathbf{x})$, where $\gamma$ and $\alpha$ are constant scalars.
For more information, please refer to Sec.~\ref{sec:control_design}.

\section{Control Design}
\label{sec:control_design}
This section will demonstrate how to design a controller that achieves the desired control objective. 
Initially, we will present the design of CLF and CBF for system \eqref{eq:dubins}. 
Subsequently, we will illustrate how to synthesize CLF and CBF to formulate a safety-critical controller.

\subsection{Design of Control Lyapunov Function}
\label{sec:dclf}
The control objective of the robot system \eqref{eq:dubins} is to navigate the robot to its destination while avoiding all obstacles. 
To achieve this, we can use CBFs for collision avoidance to ensure safety and CLF for navigation.
This section will demonstrate how to design a CLF.

Assuming that the goal position of the robot is represented by $(x_\text{g}, y_\text{g}, \theta_\text{g})$ and also denotes the coordinates of the rear axle axis.
Using $\mathbf{e} = [x_\text{p} - x_\text{g}, y_\text{p} - y_\text{g}, \theta - \theta_\text{g}]^T$ represents the difference between the current position and the goal position.
To navigate the robot to its goal position, we design a CLF as follows:
\begin{equation} 
    V(\mathbf{x}) = 
    \mathbf{e}^T
    \underbrace{
    \left[\begin{array}{ccc}
    a_1 & 0   & b_1 \\
    0   & a_2 & b_2 \\
    b_1 & b_2 & a_3
    \end{array}\right]
    }_P
    \mathbf{e},
    \label{eq:clf}
\end{equation}
where the symmetric matrix $P$ needs to be positive definite to ensure that the CLF is positive definite.
Additionally, a cross term between $x/y$ and $\theta$ is included to enhance the effectiveness of the CLF in some certain situations, such as when the orientation of both start and goal positions are identical. 
For further information, please refer to Rem.~\ref{remark1}.
\begin{remark}
\label{remark1}
Considering the CLF is designed in the form of 
\begin{equation}
    V(\mathbf{x})=(x_\text{p} - x_\text{g})^2 + (y_\text{p} - y_\text{g})^2 + (\theta - \theta_\text{g})^2 ,
\end{equation}
then we have 
\begin{equation}
    L_g V(\mathbf{x}) = \left[\begin{array}{c}
    2(x_\text{p} - x_\text{g})\cos\theta + 2(y_\text{p} - y_\text{g})\sin\theta \\
    2(\theta - \theta_\text{g})
    \end{array}\right]^T.
\end{equation}
If the start and goal positions have the same orientation, i.e., $\theta = \theta_\text{g}$, then $L_g V(\mathbf{x})$ simplifies to $[*, 0]$, indicating that the constraint \eqref{eq:kclf} depends only on the linear velocity $v$.
Consequently, the control capability of CLF is limited in this scenario, making it incapable of navigating the robot to its destination.
So it is necessary to consider the cross terms between $x/y$ and $\theta$ in $V(\mathbf{x})$.
\end{remark}

\subsection{Design of Control Barrier Function}
\label{sec:dcbf}
In this section, we will demonstrate how to design the CBFs for collision avoidance with all obstacles. 
However, it is important to note that if dynamic obstacles are present, time-varying control barrier functions (time-varying CBFs) must be used instead to avoid collision with dynamic obstacles.
Consider there is single robot sharing an open space with a set of $\mathbb{N}$ static and dynamic obstacles.
For notations, subscript $i$ is used to distinguish each obstacle, which is represented by $\text{O}_i \in \mathbb{O}=\{\text{O}_0, \text{O}_1, \dots, \text{O}_{N-1}\}$.
Assume the position of the obstacle $\text{O}_i$ is denoted by $\mathbf{z}_{\text{O}_i} (t) = (x_{\text{O}_i}(t), y_{\text{O}_i}(t))$, time-varying CBFs impose stricter constraints on the controller than regular CBFs in order to avoid collision with dynamic obstacles~\cite{igarashi2019time}, thus \eqref{eq:define_cbf} is converted to 
\begin{equation}
    \sup_{\mathbf{u} \in \mathcal{U}} [L_f h_i (\mathbf{x}, t) + L_g h_i (\mathbf{x}, t) \mathbf{u} + \frac{\partial h_i (\mathbf{x}, t)}{\partial t}] \geq -\alpha(h_i(\mathbf{x}, t)),
    \label{eq:define_tvcbf}
\end{equation}
where $\frac{\partial h_i(\mathbf{x}, t)}{\partial t} = \frac{\partial h_i(\mathbf{x}, t)}{\partial \mathbf{z}_{\text{O}_i} (t)} \frac{\partial \mathbf{z}_{\text{O}_i} (t)}{\partial t}$ shows how the position of obstacle affects the input, and the subscript $i$ of $h_i (\mathbf{x}, t)$ is used to distinguish the CBFs introduced by different obstacles.
In case of static obstacles, $\frac{\partial h_i(\mathbf{x}, t)}{\partial t} = 0$.
Therefore, the set of all controls which render the $\mathcal{C}_i$ corresponding to $h_i(\mathbf{x}, t)$ safe is equivalent to:
\begin{equation}
\begin{aligned}
    K_{\text{cbf}}^{i} (\mathbf{x}):=\{ \mathbf{u} \in \mathcal{U}, L_f h_i(\mathbf{x}, t) + L_g & h_i(\mathbf{x}, t) \mathbf{u} + \frac{\partial h_i(\mathbf{x}, t)}{\partial t} \\
    & \geq -\alpha(h_i(\mathbf{x}, t)) \}.
    \label{eq:ktvcbf}
\end{aligned}
\end{equation}

Assuming that the robot and all obstacles are in circular shapes with radii of $r_\text{r}$ and $r_{\text{O}_i}$.
Hence the safe distance between the robot and obstacle $\text{O}_i$ is defined as $r_i = r_\text{r} + r_{\text{O}_i}$.
If the distance between them exceeds this safety threshold, then safety can be guaranteed.
Therefore, we design the time-varying CBFs in the following form:
\begin{equation}
    h_i(\mathbf{x}, t) = (x_\text{p} + l\cos\theta - x_{\text{O}_i}(t))^2 + (y_\text{p} + l\sin\theta - y_{\text{O}_i}(t))^2 - r_{i}^2.
\end{equation}
And this form is also applicable to time-invariant CBFs.
By converting the form of \eqref{eq:dubins} into the format of \eqref{eq:system}, we can derive the expression for $L_g h_i(\mathbf{x}, t)$:
\begin{equation} 
    \left[\begin{array}{c}
    \dot{x}_\text{p} \\
    \dot{y}_\text{p} \\
    \dot{\theta}
    \end{array}\right] = 
    \left[\begin{array}{c}
    0 \\
    0 \\
    0
    \end{array}\right] + 
    \left[\begin{array}{cc}
    \cos \theta & 0 \\
    \sin \theta & 0 \\
    0           & 1
    \end{array}\right]
    \left[\begin{array}{c}
    v \\
    \omega
    \end{array}\right],
    \label{eq:change_form}
\end{equation}
\begin{equation}
    L_g h_i(\mathbf{x}, t) = \left[\begin{array}{c}
    2 \Delta x \cos \theta + 2 \Delta y \sin \theta \\
    -2 \Delta x l\sin \theta + 2 \Delta y l\cos \theta
    \end{array}\right]^T,
    \label{eq:lg_h}
\end{equation}
where
\begin{equation}
    \left[\begin{array}{c}
        \Delta x \\
        \Delta y
    \end{array}\right] = 
    \left[\begin{array}{c}
        x_\text{p} + l\cos\theta - x_{\text{O}_i}(t) \\
        y_\text{p} + l\sin\theta - y_{\text{O}_i}(t)
    \end{array}\right].
\end{equation}
It can be verified that if $L_g h_i(\mathbf{x}, t)$ to be a zero matrix, we have the following case:
$\Delta x \sin\theta = \Delta y \cos\theta$, without loss of generality, we assume $\cos\theta \not= 0$.
Then we can simplify this equation to get $\Delta y = \frac{\sin\theta}{\cos\theta}\Delta x$, next we can get $\Delta x = \Delta  y =0$.
However, if both $\Delta x$ and $\Delta y$ equal to zero, it would mean that the robot is already inside the obstacle, which is impossible.
Therefore, $L_g h_i(\mathbf{x}, t)$ is always non-zero to ensure $h_i(\mathbf{x}, t)$ to be a valid CBF.
\begin{remark}
\label{remark2}
If $(x_\text{p}, y_\text{p})$ denotes the coordinates of center of the robot, then the CBF corresponding to the obstacle $\text{O}_i$ can be expressed as 
\begin{equation}
    h_i(\mathbf{x}, t) = (x_\text{p} - x_{\text{O}_i}(t))^2 + (y_\text{p} - y_{\text{O}_i}(t))^2 - r_{i}^2.
    \label{eq:time_invariant_cbf}
\end{equation}
In this case, $L_g h_i(\mathbf{x}, t)$ is equal to $[*, 0]$, which means that \eqref{eq:ktvcbf} only relies on $v$.
This limits the control performance of CBF by causing it to lose control of steering $w$, which may result in deadlock in some cases.
\end{remark}

\begin{remark}
\label{remark3}
If we choose to utilize HOCBF, it is necessary to fix the linear velocity, then the system \eqref{eq:dubins} can be converted into the format of \eqref{eq:system} as follow:
\begin{equation} 
    \left[\begin{array}{c}
    \dot{x}_\text{p} \\
    \dot{y}_\text{p} \\
    \dot{\theta}
    \end{array}\right] = 
    \left[\begin{array}{c}
    v \cos \theta \\
    v \sin \theta\\
    0
    \end{array}\right] + 
    \left[\begin{array}{c}
     0 \\
     0 \\
     1
    \end{array}\right]
    w.
    \label{eq:fixed_form}
\end{equation}
With the CBF form as \eqref{eq:time_invariant_cbf}, we can get
\begin{equation}
    L_f h_i(\mathbf{x}, t) = 2(x_\text{p} - x_{\text{O}_i}(t))v \cos \theta + 2(y_\text{p} - y_{\text{O}_i}(t))v \sin \theta
\end{equation} 
and $L_g h_i(\mathbf{x}, t) = 0$.
Thus, we can get 
\begin{equation}
    L_g L_f h_i(\mathbf{x}, t) = -2(x_\text{p} - x_{\text{O}_i}(t))v \sin \theta + 2(y_\text{p} - y_{\text{O}_i}(t))v \cos \theta.
\end{equation}
Thus the controller can control steering $w$ to avoid collision with obstacles, but the control performance is limited due to the fixed linear velocity.
\end{remark}

\subsection{Controller Synthesis}
Since the constraints \eqref{eq:kclf} and \eqref{eq:ktvcbf} in $\mathbf{u}$ have an affine form, real-time solutions can be generated. 
Therefore, we propose a Quadratic Program (QP) based formation of the safety-critical controller that combines CBFs for safety and CLF for stability as follows:

\noindent\rule{\columnwidth}{0.8pt}
\textbf{CLF-CBF-QP:}
\begin{subequations}
\begin{align}
    \min_{(\mathbf{u}, \delta) \in \mathbb{R}^{m + 1}} & \frac{1}{2}\mathbf{u}^T H \mathbf{u} + p\delta ^ 2 + (\mathbf{u} - \mathbf{u}_\text{pre})^T Q (\mathbf{u} - \mathbf{u}_\text{pre})
    \label{eq:problem} \\
    \text{s.t.} ~& L_f V(\mathbf{x}) + L_g V(\mathbf{x}) \mathbf{u} + \gamma(V(\mathbf{x})) \leq \delta, \label{eq:cons_clf} \\ \notag
    & L_f h_i(\mathbf{x}, t) + L_g h_i(\mathbf{x}, t) \mathbf{u} + \frac{\partial h_i(\mathbf{x}, t)}{\partial t} \\ 
    & + \alpha(h_i(\mathbf{x}, t)) \geq 0, i=0, 1, \dots, N-1,  \label{eq:cons_cbf} \\
    & \mathbf{u} \in \mathcal{U}, \label{eq:cons_u} \\
    & \delta \in \mathbb{R}. \label{eq:cons_delta} 
\end{align}
\label{eq:optimal_problem}
\end{subequations}
\noindent\rule{\columnwidth}{0.4pt}
\noindent
where $H$ and $Q$ are any positive definite matrices, $p>0$ is the weight coefficient of the relaxation variable, and $\mathbf{u}_\text{pre}$ is the control value at the previous moment.
For the objective function \eqref{eq:problem} in this optimization problem, we divided it into three parts:
the first part requires $\mathbf{u}$ to be as small as possible, the second part is the additional quadratic cost of the relaxation variable, and the third part requires $\mathbf{u}$ to change smoothly.
The constraint \eqref{eq:cons_clf} of CLF is relaxed by the relaxation variable $\delta$, which means that when the constraint \eqref{eq:cons_clf} conflicts with constraints \eqref{eq:cons_cbf}, the controller must relax the condition on stability to guarantee safety.
Constraints \eqref{eq:cons_cbf} imply that the optimal control problem must satisfy all safety constraints to guarantee system safety.
The constraint \eqref{eq:cons_u} requires that the control $\mathbf{u}$ must be within the permissible range.
\section{Numerical validation}
\label{sec:experiment}
\begin{table}
\caption{Setup of the simulation parameter}
\label{tab:simulation_para}
\centering
\begin{tabular}{l|l|l}
\hline
Notation & Meaning & Value     \\ \hline
$r_\text{r}$ & Robot's radius & $0.25 \, \si[per-mode=symbol]{\metre}$ \\
$l$ & Distance between rear axle axis and the center & $0.15 \, \si[per-mode=symbol]{\metre}$ \\
$v_{\text{max}}$ & Robot's maximum linear velocity & $2.0 \, \si[per-mode=symbol]{\metre\per\second}$ \\
$w_{\text{max}}$ & Robot's maximum angular velocity & $1.5 \, \si[per-mode=symbol]{\radian\per\second}$ \\
$p$ & Weight coefficient of the relaxation variable & 1000 \\
$\alpha$ & Class $\mathcal{K}$ function for control barrier functions & 1.5 \\
$\gamma$ & Class $\mathcal{K}$ function for control Lyapunov function & 0.5 \\ \hline     \end{tabular}%
\end{table}

\begin{figure}
    \centering
    \begin{subfigure}{0.49\linewidth}
        \centering
        \includegraphics[width=0.95\linewidth]{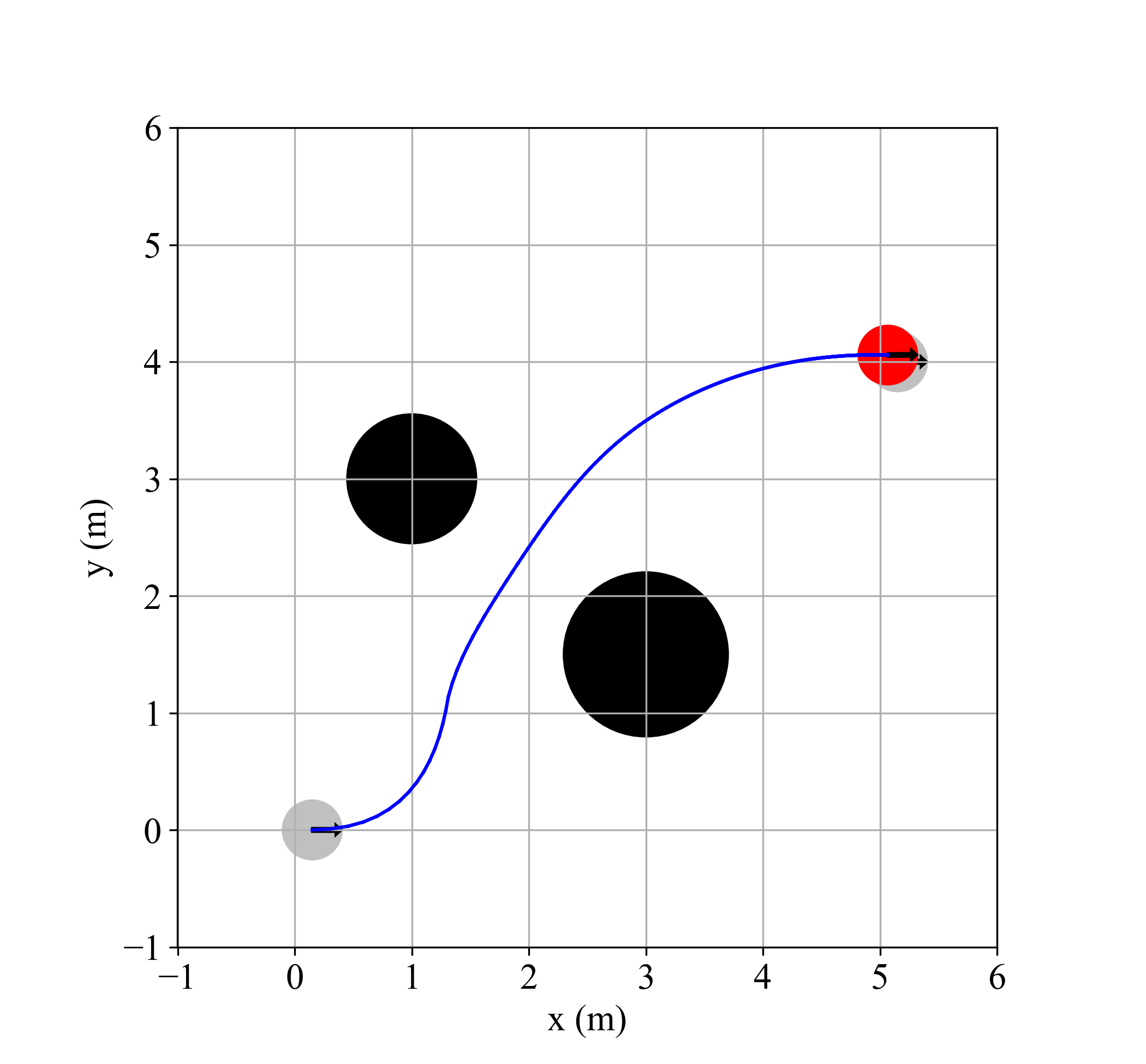}
        \caption{With our proposed method}
        \label{subfig:obstacle_case1}
    \end{subfigure}
    \centering
    \begin{subfigure}{0.49\linewidth}
        \centering
        \includegraphics[width=0.95\linewidth]{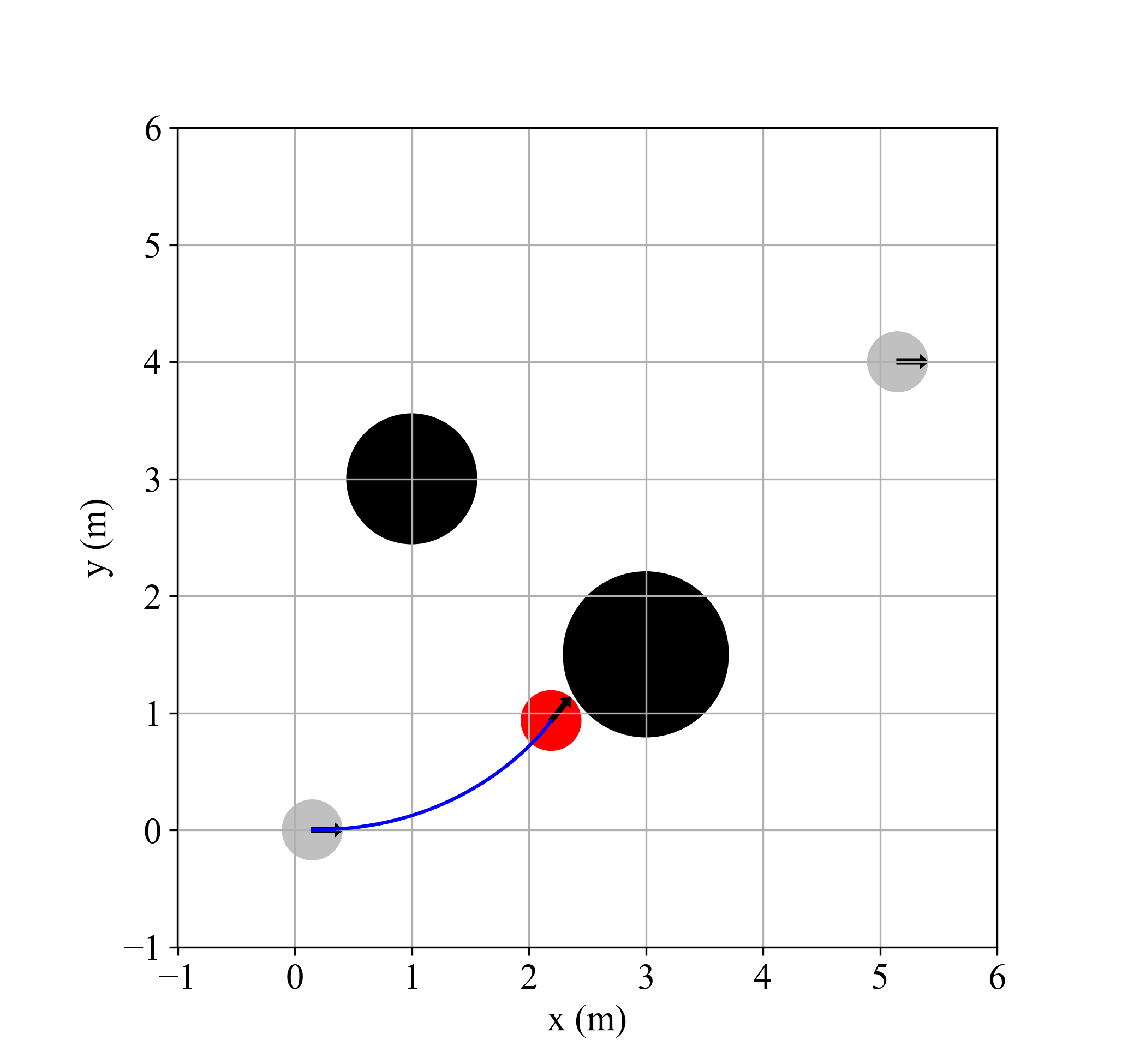}
        \caption{With traditional method}
        \label{subfig:obstacle_case2}
    \end{subfigure}
    \caption{Simulation results of the navigation progress in the presence of static obstacles. The black circles signify these static obstacles, while the red circle represents the robot and the silver circles denote its initial and goal positions. Additionally, the blue line illustrates the robot's previous trajectory.} 
    \label{fig:static_obstacle}
\end{figure}

\begin{figure*}
    \centering
    \begin{subfigure}{0.32\linewidth}
        \centering
        \includegraphics[width=0.95\linewidth]{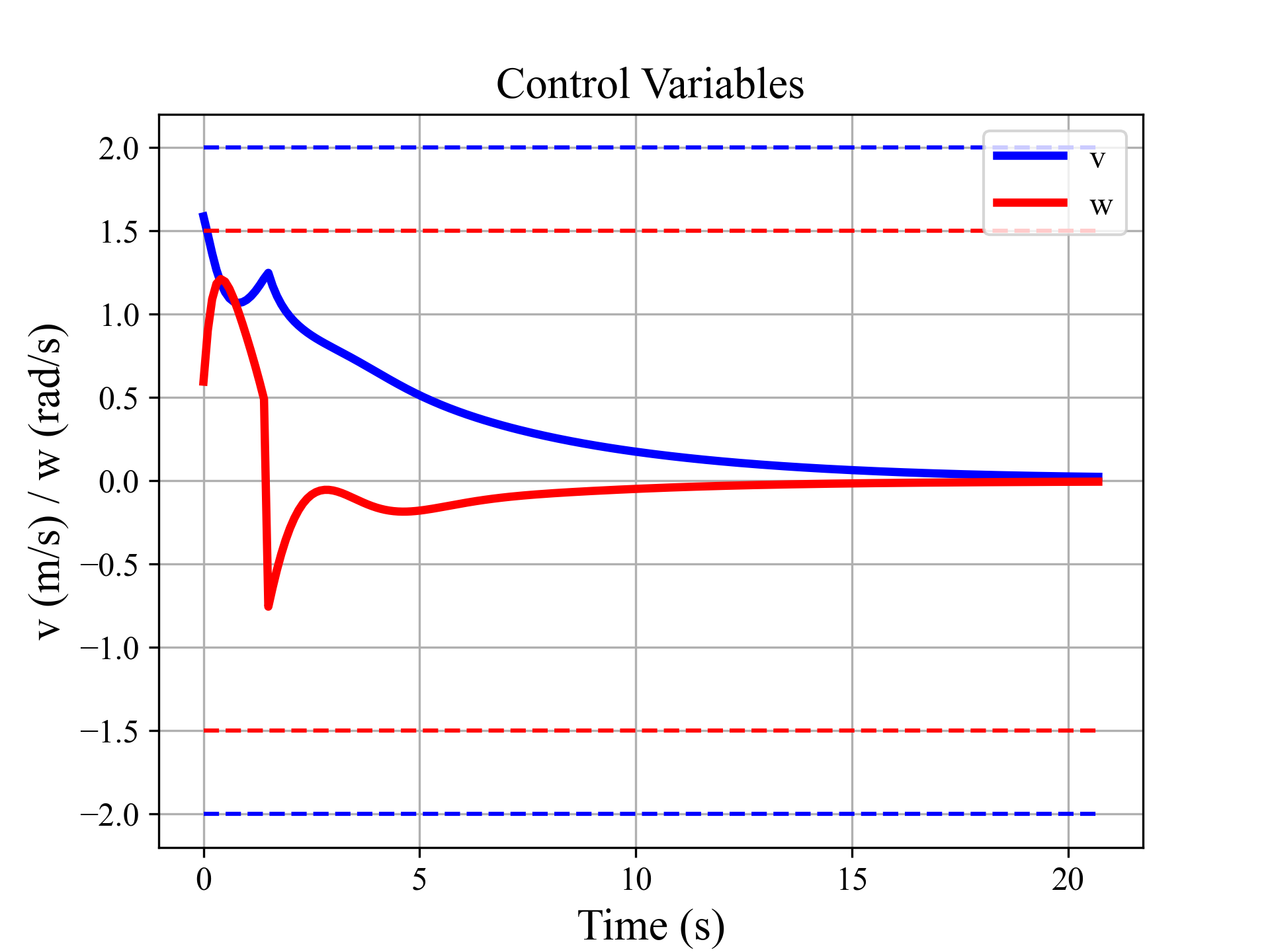}
        \caption{Changes in control variables}
        \label{subfig:static_control}
    \end{subfigure}
    \centering
    \begin{subfigure}{0.32\linewidth}
        \centering
        \includegraphics[width=0.95\linewidth]{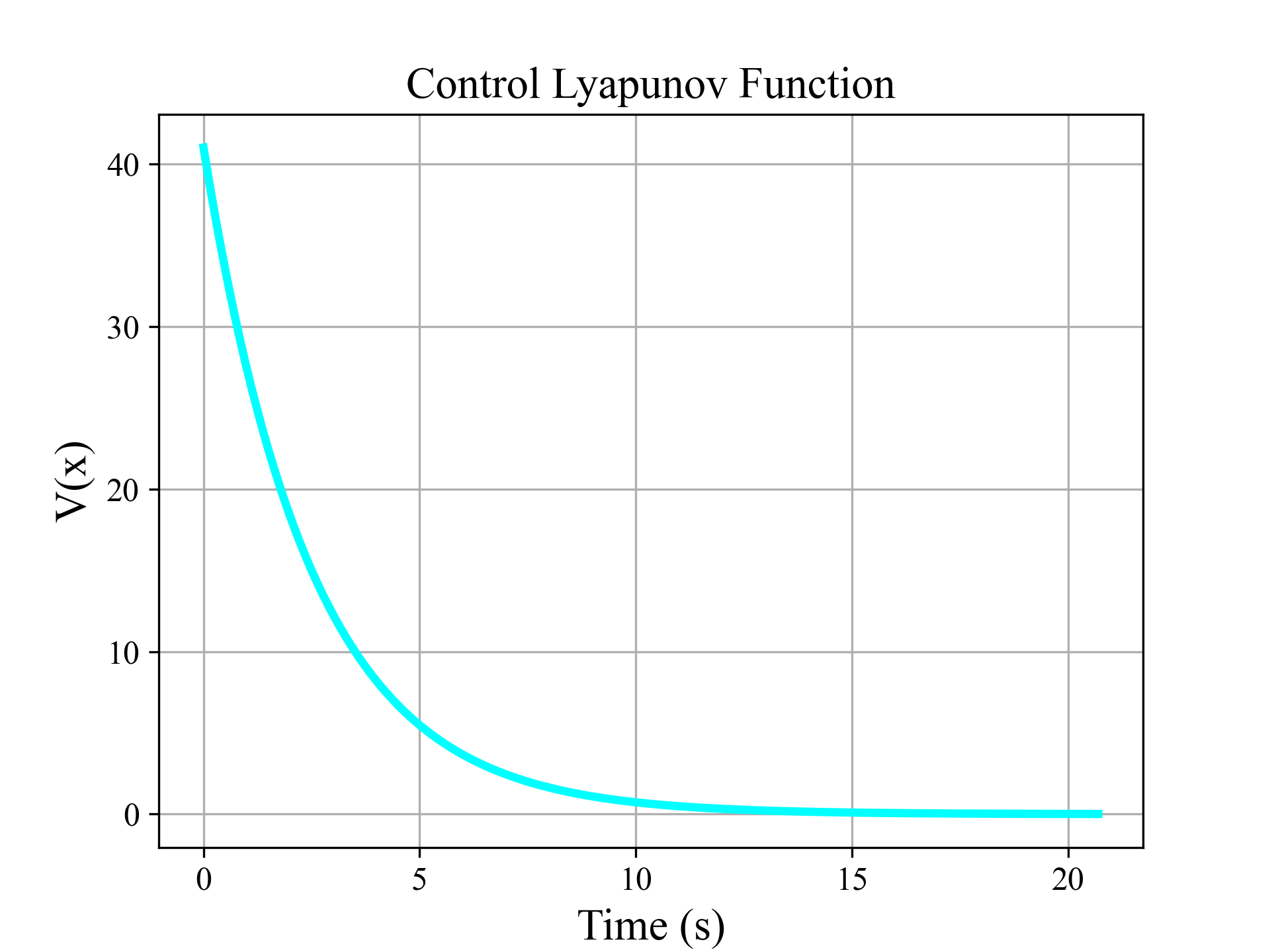}
        \caption{Changes in control Lyapunov function}
        \label{subfig:static_clf}
    \end{subfigure}
    \centering
    \begin{subfigure}{0.32\linewidth}
        \centering
        \includegraphics[width=0.95\linewidth]{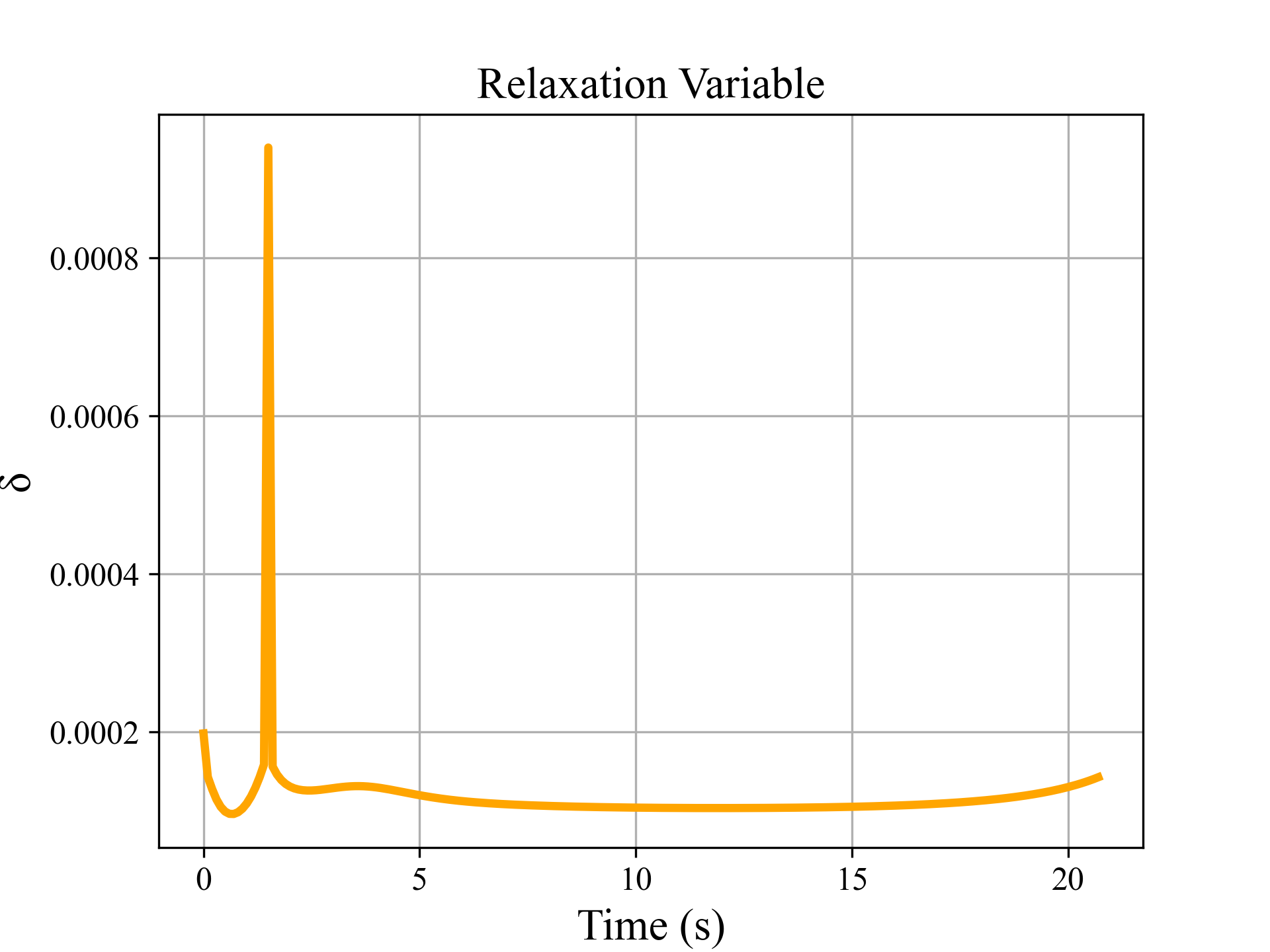}
        \caption{Changes in relaxation variable}
        \label{subfig:static_slack}
    \end{subfigure}
    \caption{Changes over time in control variables, control Lyapunov function value and relaxation variable during navigation with static obstacles.}  
    \label{fig:static_variable}
\end{figure*}

In this section, we validate the effectiveness of our proposed work through numerical validation. 
The simulations are conducted on an Ubuntu Laptop with Intel Core i9-13900HX processor using Python for all computations. We solved the online QP using IPOPT. 
The time step $\Delta t$ of simulation is set as $0.1 \, \si[per-mode=symbol]{\second}$ and the maximum loop time $t_{\text{max}}$ is set as $30\, \si[per-mode=symbol]{\second}$.
And $-\mathbf{u}_\text{min} = \mathbf{u}_\text{max} = [v_\text{max}, w_\text{max}]^T$, where $v_\text{max}$  and $w_\text{max}$ represent the maximum of linear velocity and angular velocity. 
We have presented the parameters for both the robot model and the CLF-CBF-QP controller in Tab.~\ref{tab:simulation_para}.
In the following we will present the simulation results of various scenarios along with detailed discussions.

\subsection{Effectiveness of Control Barrier Functions for Static Obstacles}
In this section, we demonstrate the effectiveness of time-invariant CBFs in achieving collision avoidance with static obstacles while navigating a robot to its goal position using CLF.
The initial and goal positions of robot are $(0.0 \, \si[per-mode=symbol]{\metre}, 0.0 \, \si[per-mode=symbol]{\metre}, 0.0 \, \si[per-mode=symbol]{\radian})$ and $(5.0 \, \si[per-mode=symbol]{\metre}, 4.0 \, \si[per-mode=symbol]{\metre}, 0.0 \, \si[per-mode=symbol]{\radian})$, respectively, and there are two static obstacles in the environment.
Our approach can navigate the robot to its destination while avoiding collisions with all these obstacles, as shown in Fig.~\ref{subfig:obstacle_case1}.
Furthermore, if we use the traditional method which uses \eqref{eq:time_invariant_cbf} as the CBFs constraint, then the robot could not adjust its angular velocity to avoid collision with obstacles, it will collide with obstacles, as shown in Fig.~\ref{subfig:obstacle_case2}.

In Fig.~\ref{fig:static_variable}, we present the changes in control variable, CLF value, and relaxation variable over time. 
We can find that the control variable is within the allowable range, as indicated by dashed lines of different colors shown in Fig.~\ref{subfig:static_control}.
The smoothness of the changes in control variable is due to the design of our objective function.
Additionally, we observe an exponential decrease in CLF value, indicating that the robot moves towards its destination.
Owing to a large weight coefficient for relaxation variable, it remains negligible throughout the process, as shown in Fig.~\ref{subfig:static_slack}.
Our approach has an average computation time of $0.0069 \,\si[per-mode=symbol]{\second}$ making it suitable for real-time applications.

\subsection{Effectiveness of Control Barrier Functions for Dynamic Obstacles}
\begin{figure}
    \centering
    \begin{subfigure}{0.49\linewidth}
        \centering
        \includegraphics[width=0.98\linewidth]{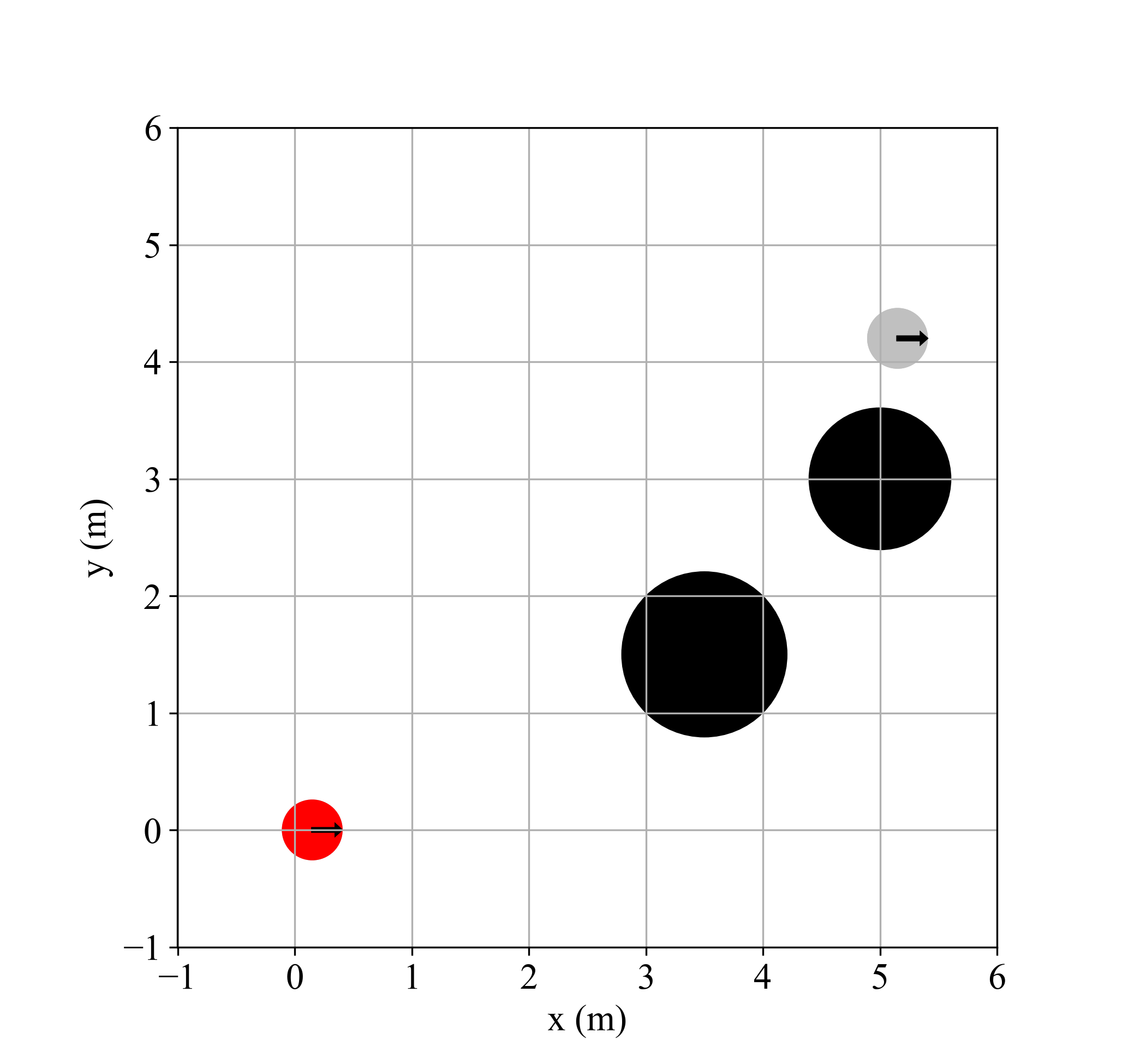}
        \caption{$t = 0.0 \, \si[per-mode=symbol]{\second}$}
        \label{subfig:dynamic_case1}
    \end{subfigure}
    \centering
    \begin{subfigure}{0.49\linewidth}
        \centering
        \includegraphics[width=0.98\linewidth]{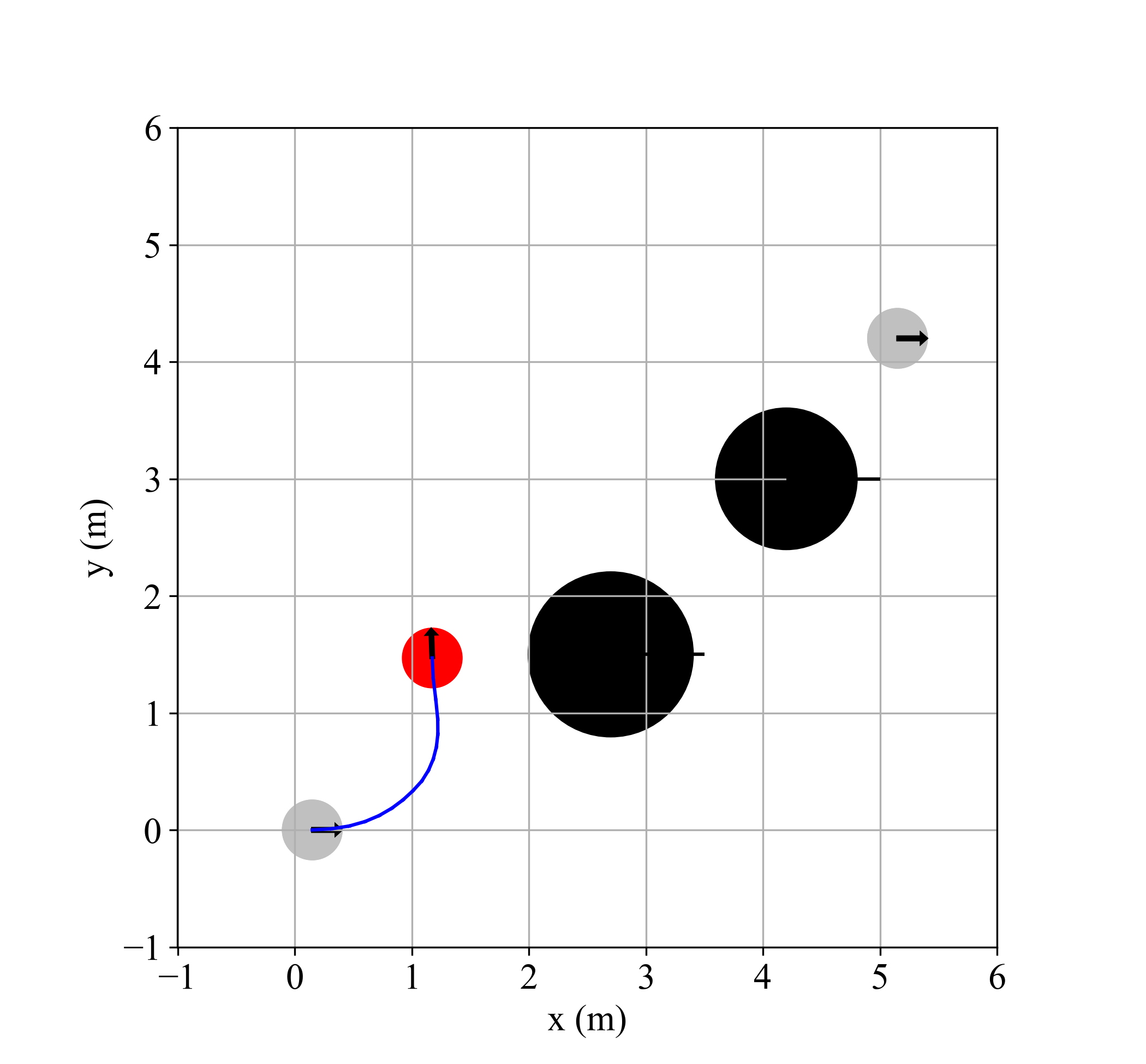}
        \caption{$t = 1.6 \, \si[per-mode=symbol]{\second}$}
        \label{subfig:dynamic_case2}
    \end{subfigure}

    \centering
    \begin{subfigure}{0.49\linewidth}
        \centering
        \includegraphics[width=0.98\linewidth]{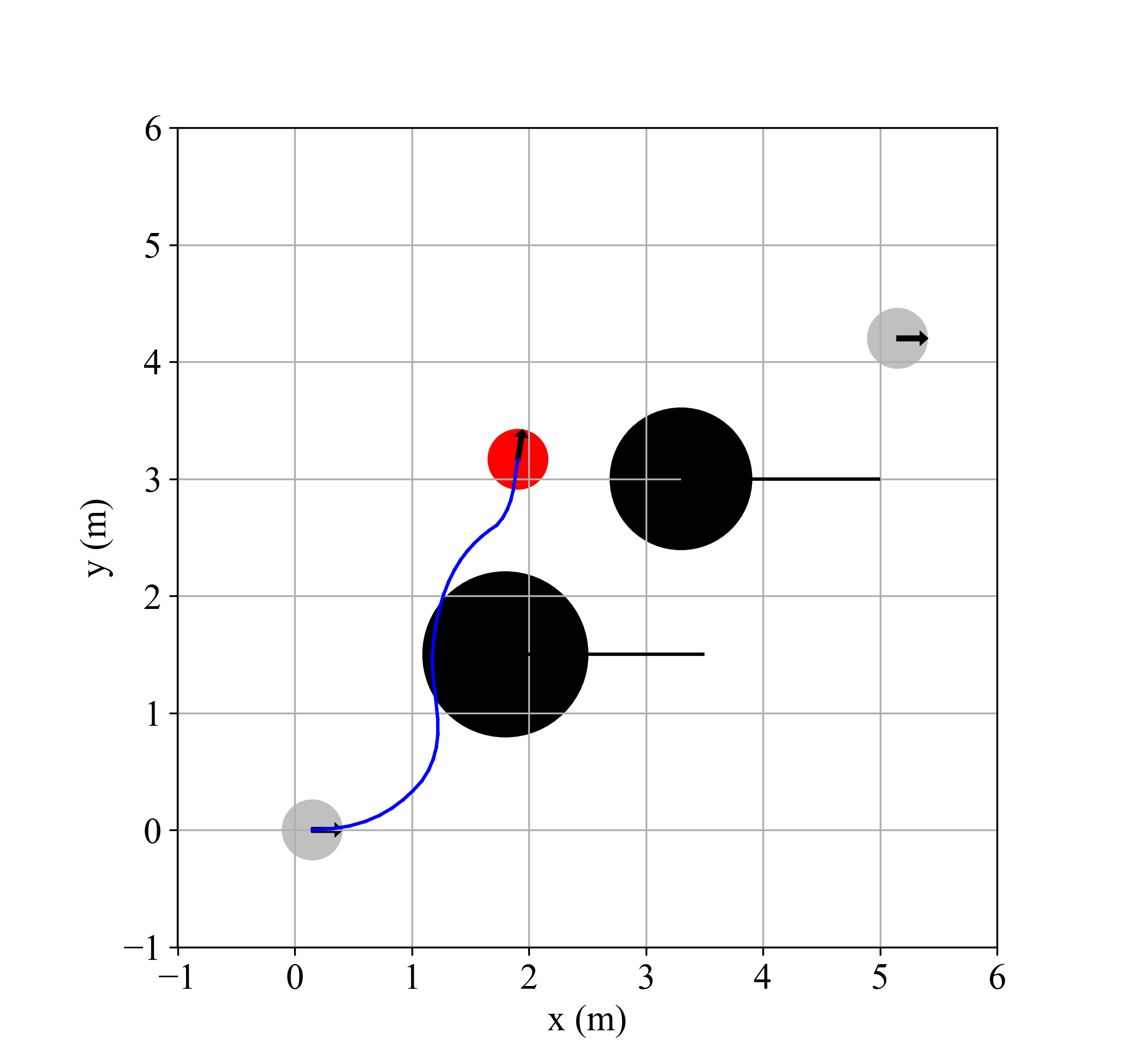}
        \caption{$t = 3.4 \, \si[per-mode=symbol]{\second}$}
        \label{subfig:dynamic_case3}
    \end{subfigure}
    \centering
    \begin{subfigure}{0.49\linewidth}
        \centering
        \includegraphics[width=0.98\linewidth]{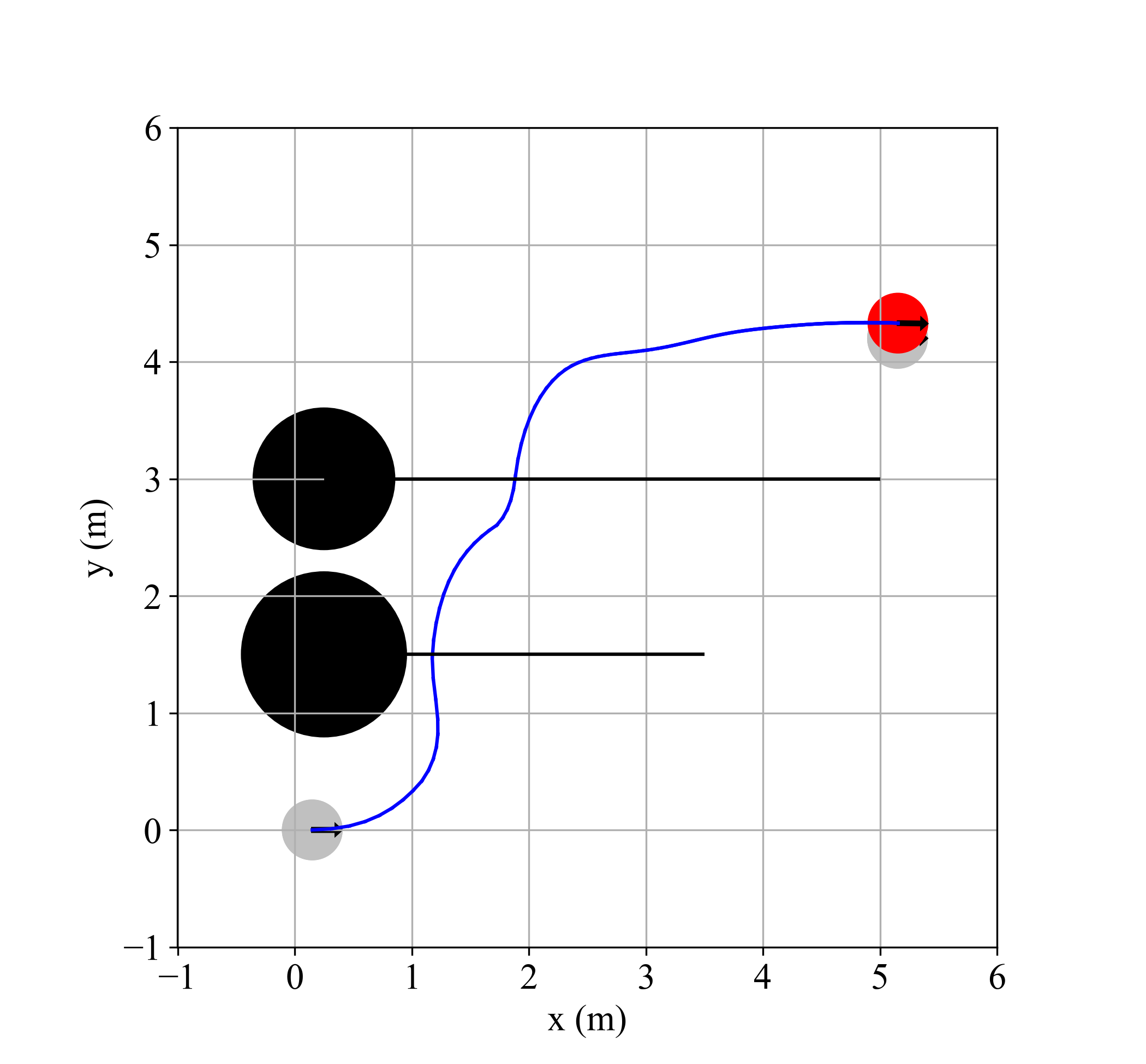}
        \caption{$t = 20.0 \, \si[per-mode=symbol]{\second}$}
        \label{subfig:dynamic_case4}
    \end{subfigure}
    \caption{Simulation results of the navigation progress in the presence of dynamic obstacles. The red circle represents the robot, while black circles denote dynamic obstacles. Additionally, silver circles indicate the initial and goal positions of the robot. The blue line illustrates the past trajectory of the robot, whereas black lines represent those of dynamic obstacles.} 
    \label{fig:dynamic_obstacle}
\end{figure}

\begin{figure}
    \centering
    \begin{subfigure}{0.49\linewidth}
        \centering
        \includegraphics[width=0.95\linewidth]{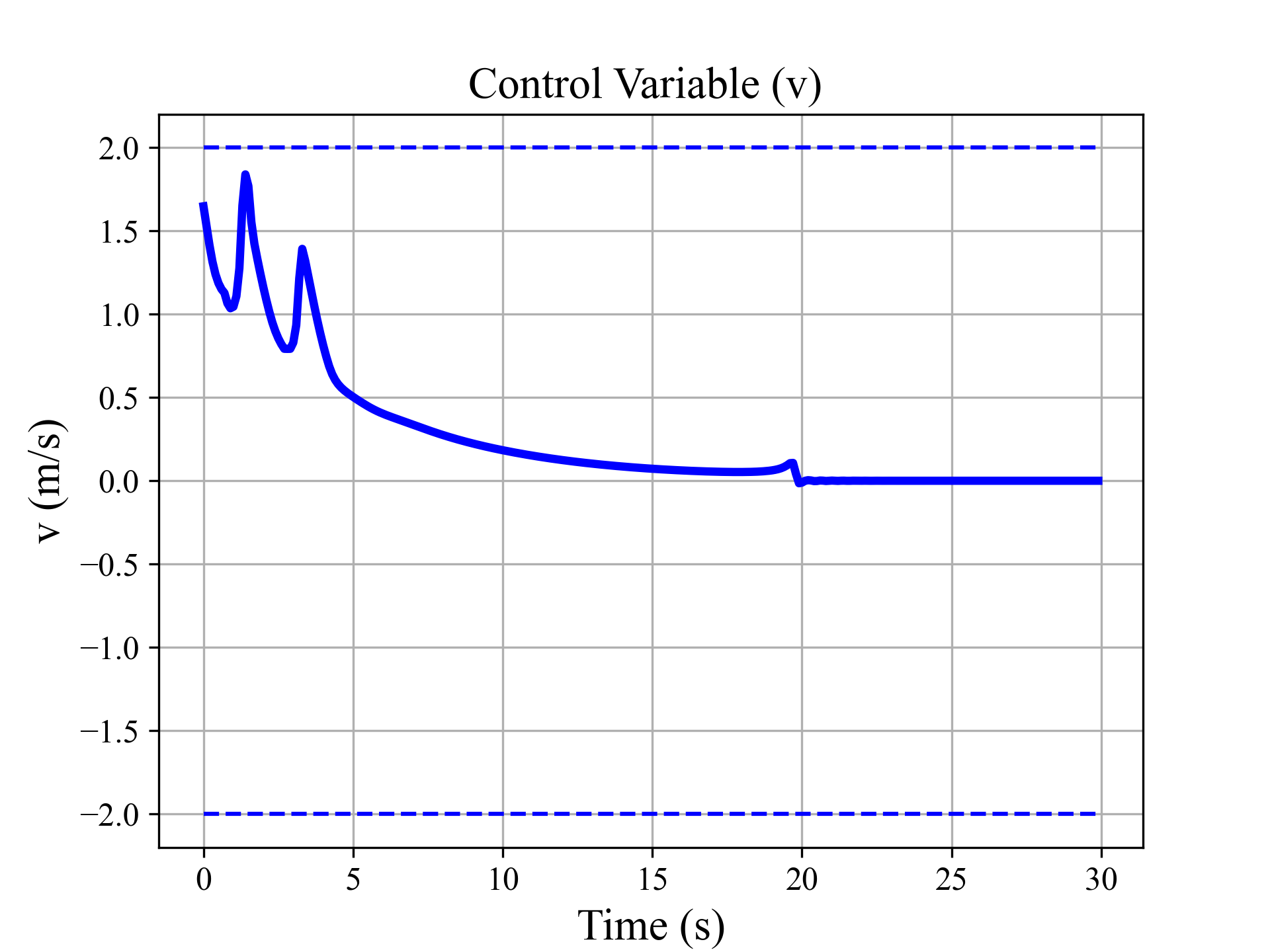}
        \caption{Changes in control variable $v$}
        \label{subfig:dynamic_control_v}
    \end{subfigure}
    \centering
    \begin{subfigure}{0.49\linewidth}
        \centering
        \includegraphics[width=0.95\linewidth]{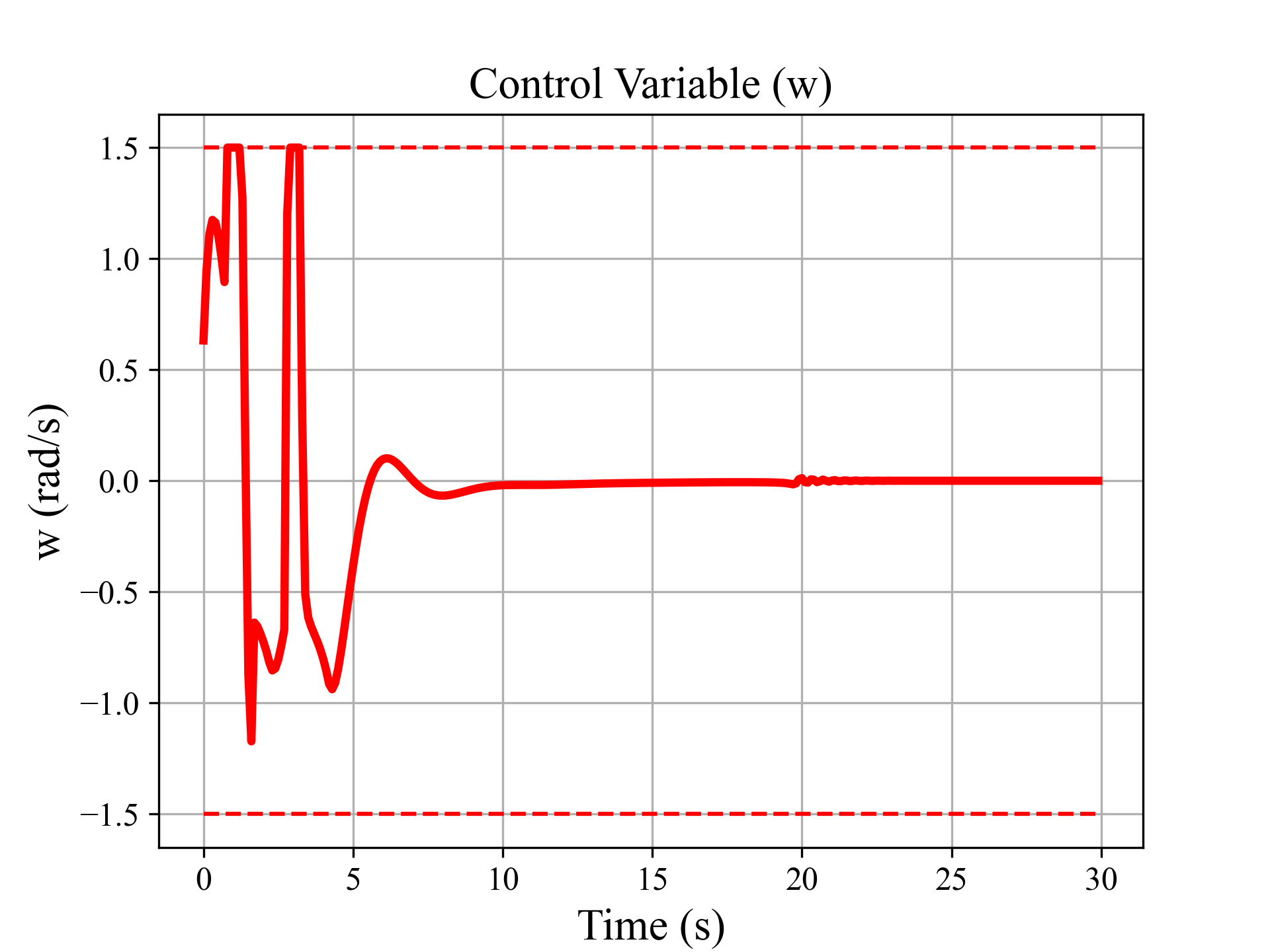}
        \caption{Changes in control variable $w$}
        \label{subfig:dynamic_control_w}
    \end{subfigure}
    \\
    \centering
    \begin{subfigure}{0.49\linewidth}
        \centering
        \includegraphics[width=0.95\linewidth]{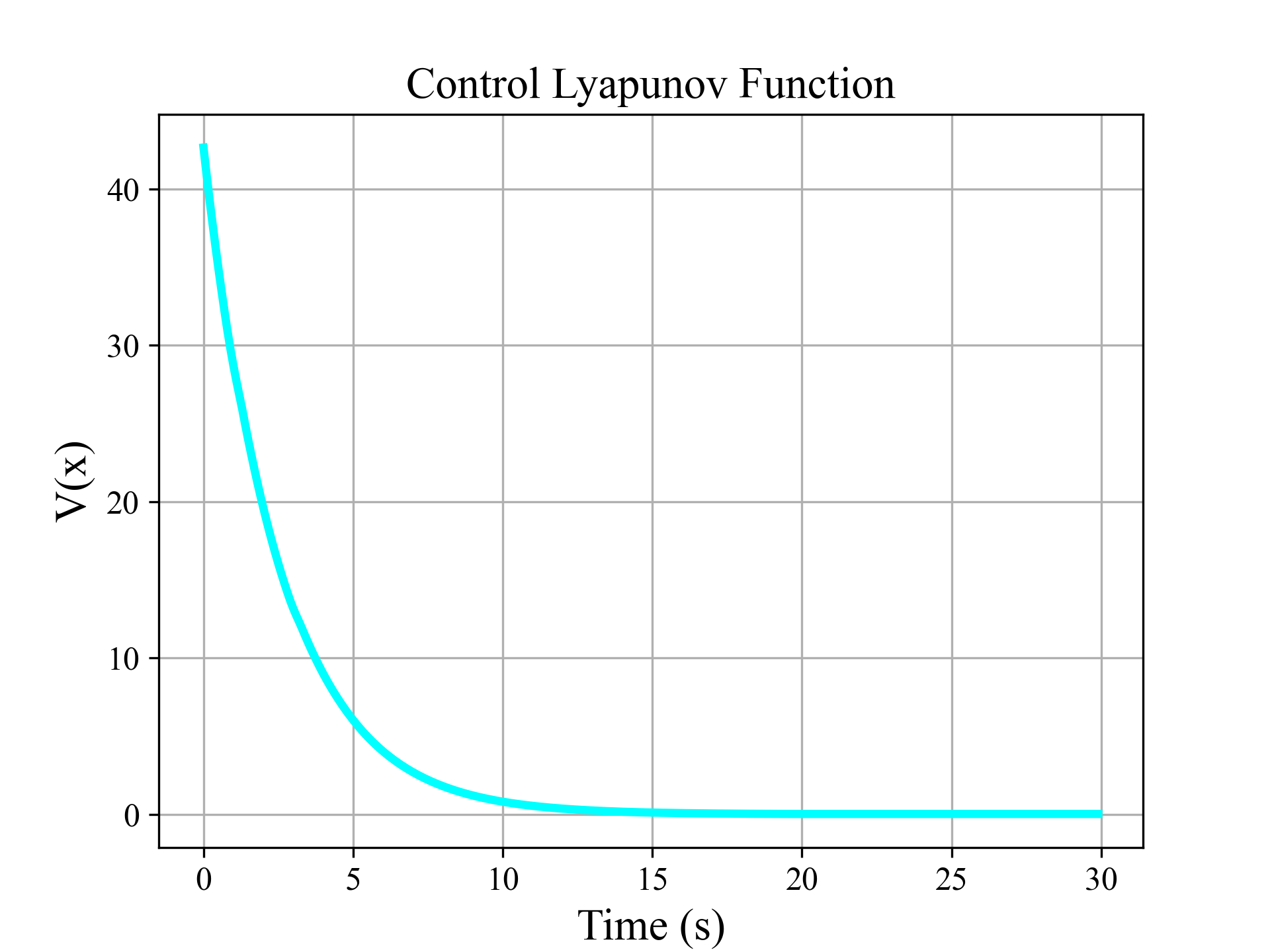}
        \caption{Changes in CLF}
        \label{subfig:dynamic_clf}
    \end{subfigure}
    \centering
    \begin{subfigure}{0.49\linewidth}
        \centering
        \includegraphics[width=0.95\linewidth]{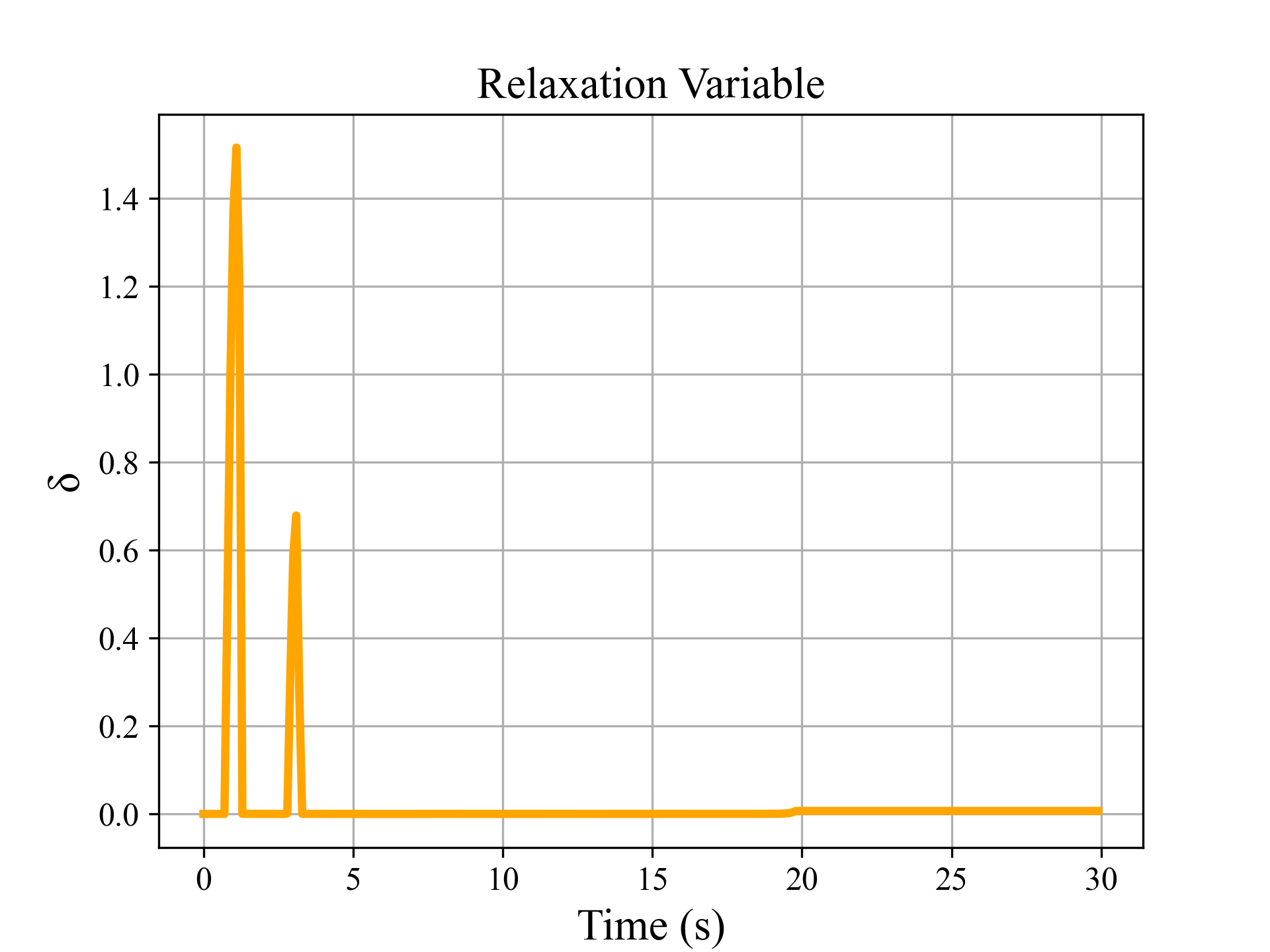}
        \caption{Changes in relaxation variable}
        \label{subfig:dynamic_slack}
    \end{subfigure}
    \caption{Changes over time in control variables, control Lyapunov function (CLF) value and relaxation variable during navigation with dynamic obstacles.}
    \label{fig:dynamic_variable}
\end{figure}

This section validates the effectiveness of time-varying CBFs in achieving collision avoidance with dynamic obstacles while navigating a robot to its goal position using CLF. 
The robot's initial and goal positions are set at $(0.0 \, \si[per-mode=symbol]{\metre}, 0.0 \, \si[per-mode=symbol]{\metre}, 0.0 \, \si[per-mode=symbol]{\radian})$ and $(5.0 \, \si[per-mode=symbol]{\metre}, 4.2 \, \si[per-mode=symbol]{\metre}, 0.0 \, \si[per-mode=symbol]{\radian})$, respectively.
There are two dynamic obstacles in the environment: one travels from $(3.5 \, \si[per-mode=symbol]{\metre}, 1.5 \, \si[per-mode=symbol]{\metre})$ to $(0.3 \, \si[per-mode=symbol]{\metre}, 1.5 \, \si[per-mode=symbol]{\metre})$ at a speed of $0.5 \, \si[per-mode=symbol]{\metre\per\second}$, the other travels from $(5.0 \, \si[per-mode=symbol]{\metre}, 3.0 \, \si[per-mode=symbol]{\metre})$ to $(0.3 \, \si[per-mode=symbol]{\metre}, 3.0 \, \si[per-mode=symbol]{\metre})$ at a speed of $0.5 \, \si[per-mode=symbol]{\metre\per\second}$.
Our approach successfully navigates the robot to its destination while avoiding collisions with all dynamic obstacles, as shown in Fig.~\ref{fig:dynamic_obstacle}.
We show four different moments during the navigation process and in Fig.~\ref{subfig:dynamic_case2} and Fig.~\ref{subfig:dynamic_case3} we can observe that the robot adjusts its velocity to avoid collision with dynamic obstacles.

In Fig.~\ref{fig:dynamic_variable}, we present the changes in control values, CLF value, and relaxation variable over time while navigating with dynamic obstacles. 
The control variables in Fig.~\ref{subfig:dynamic_control_v} and Fig.~\ref{subfig:dynamic_control_w} exhibit significant changes, particularly when steering to avoid collision with dynamic obstacles or to move towards the destination, demonstrating the effectiveness of our proposed controller.
Additionally, Fig.~\ref{subfig:dynamic_slack} shows two peaks in the relaxation variable.
As previously mentioned, the constraint \eqref{eq:cons_clf} of CLF is relaxed using a relaxation variable to satisfy constraints \eqref{eq:cons_cbf} of CBFs when these two conflicts.
In this case, since the robot needs to avoid two dynamic obstacles sequentially, there are two peaks in the changes of relaxation variable over time.
Finally, our approach has an average computation time of $0.0066 \,\si[per-mode=symbol]{\second}$ for the entire process which indicates that it can work in real-time for dynamic collision avoidance purposes. 

\section{Conclusions}
\label{sec:conclusion}
In this section, we present a safety-critical controller for the unicycle model robot. 
Our approach unifies a control Lyapunov function (CLF) and time-varying control barrier functions (CBFs) through a quadratic program (CLF-CBF-QP) to achieve navigation and collision avoidance with both static and dynamic obstacles.
We conducted several numerical validations to demonstrate the effectiveness of our proposed approach, and the results verify that our approach successfully navigates the robot to its goal position while avoiding collisions with both static and dynamic obstacles.
Our proposed work has certain limitations as it primarily focuses on circular obstacles and robots. 
When applied to non-circular items, it may prove to be ineffective or approximate these items as circles with a significant error.
To address this issue, future research will focus on extending our proposed approach to accurately handle non-circular items, such as polytopes.
Furthermore, extending our approach to multi-robot systems will be an important direction for future research.

{
\bibliographystyle{IEEEtran}
\bibliography{reference}
}

\end{document}